\documentclass[conference]{IEEEtran}
\IEEEoverridecommandlockouts
\usepackage{cite}
\usepackage{amsmath,amssymb,amsfonts}
\usepackage{algorithmic}
\usepackage{graphicx}
\usepackage{textcomp, booktabs}
\usepackage{xcolor}
\def\BibTeX{{\rm B\kern-.05em{\sc i\kern-.025em b}\kern-.08em
    T\kern-.1667em\lower.7ex\hbox{E}\kern-.125emX}}

\def\cE{{\cal E}}

\def\cH{{\cal H}}
\def\cK{{\cal K}}
\def\cX{{\cal X}}

\def\rT{{\rm T}}

\def\uR{{\mathbb R}}

\def\be{ \begin{equation} }
\def\ee{ \end{equation} }
\def\bea{ \begin{eqnarray} }
\def\eea{ \end{eqnarray} }

\def\bx{{\bf x}}

\def\bg{{\bf g}}

\def\bs{{\bf s}}

\def\bh{{\bf h}}

\def\bz{{\bf z}}

\def\bF{{\bf F}}

\def\bK{{\bf K}}

\def\bS{{\bf S}}

\def\bW{{\bf W}}

\def\b0{{\bf 0}}

\def\cL{{\cal L}}
\def\cD{{\cal D}}

\def\cI{{\cal I}}
\def\cN{{\cal N}}

\def\cU{{\cal U}}

\ifCLASSOPTIONonecolumn
\else

\fi

\usepackage{geometry}
\begin{document}

\title{Koopman-based Prediction of Connectivity for
Flying Ad Hoc Networks\\
}

\author{
    \IEEEauthorblockN{
         Sivaram Krishnan\IEEEauthorrefmark{1}, 
         Jinho Choi\IEEEauthorrefmark{1}, 
         Jihong Park\IEEEauthorrefmark{2}, 
         Gregory Sherman\IEEEauthorrefmark{3}, 
         and Benjamin Campbell\IEEEauthorrefmark{3}
     }
     \IEEEauthorblockA{
         \IEEEauthorrefmark{1}School of Electrical and Mechanical Engineering, The University of Adelaide, Adelaide, Australia \\
         Email: \{sivaram.krishnan, jinho.choi\}@adelaide.edu.au
     }
     \IEEEauthorblockA{
         \IEEEauthorrefmark{2}Information Systems Technology and Design Pillar, Singapore University of Technology and Design, Singapore \\
         Email: jihong\_park@sutd.edu.sg
     }
     \IEEEauthorblockA{
         \IEEEauthorrefmark{3}Mission Autonomy, Platforms Division, Defence Science and Technology Group, Australia    }
}
\maketitle

\begin{abstract}

The application of machine learning (ML) to communication systems is expected to play a pivotal role in future artificial intelligence (AI)-based next-generation wireless networks. While most existing works focus on ML techniques for static wireless environments, they often face limitations when applied to highly dynamic environments, such as flying ad hoc networks (FANETs). This paper explores the use of data-driven Koopman approaches to address these challenges. Specifically, we investigate how these approaches can model UAV trajectory dynamics within FANETs, enabling more accurate predictions and improved network performance. By leveraging Koopman operator theory, we propose two possible approaches—centralized and distributed—to efficiently address the challenges posed by the constantly changing topology of FANETs. To demonstrate this, we consider a FANET performing surveillance with UAVs following pre-determined trajectories and predict signal-to-interference-plus-noise ratios (SINRs) to ensure reliable communication between UAVs. Our results show that these approaches can accurately predict connectivity and isolation events that lead to modelled communication outages. This capability could help UAVs schedule their transmissions based on these predictions.
\end{abstract}


\section{Introduction}

Wireless ad hoc networks are decentralized wireless networks where nodes communicate directly with each other without relying on fixed infrastructure such as base stations or access points \cite{Toh101}. These networks are highly flexible and can be rapidly deployed in scenarios where traditional communication infrastructure is unavailable or impractical, such as disaster recovery, military operations, and remote monitoring. When nodes are mobile, the network evolves into a mobile ad hoc network (MANET) \cite{Loo16e}, which introduces additional challenges. These challenges include dynamic topology changes due to node mobility, frequent link breakages, and the need for efficient routing protocols to ensure reliable communication.

An extension of MANETs is the flying ad hoc network (FANET) \cite{Khan17}, where unmanned aerial vehicles (UAVs) serve as mobile nodes to form a network. FANETs offer unique advantages such as enhanced coverage, flexibility, and scalability, making them particularly valuable for applications like aerial surveillance, environmental monitoring, and disaster response. Furthermore, FANETs can be integrated with terrestrial networks to extend coverage and improve connectivity \cite{Nemati2022-sb}. 

There are various challenges in FANETs arising from the high mobility of nodes, which can result in frequent topology changes and connectivity issues. While UAV locations can be dynamically controlled to optimize network performance \cite{Cicek19} \cite{He24}, an alternative approach involves UAVs \emph{following pre-determined trajectories}. This approach resembles the operation of low Earth orbit (LEO) satellites, which move along predefined orbits to form a network.
In such cases, their connectivity is expected to exhibit periodic and deterministic patterns. This predictable mobility can be exploited to enhance network reliability and simplify the design of routing protocols. 
Consequently, accurate modeling UAV dynamics becomes essential for effectively predicting mobility.

The trajectory of a UAV can be modeled as a nonlinear dynamical system, as their movements often involve complex and nonlinear behaviors. Therefore, employing effective modeling techniques for nonlinear dynamical systems is crucial. Among these techniques, approaches based on Koopman operator theory \cite{Koopman31} are particularly noteworthy. The Koopman operator enables the linearization of nonlinear systems by transforming the original system into a higher-dimensional space, where the dynamics can be approximated as linear. This transformation facilitates the use of linear analysis tools to study and predict the system's behavior \cite{Budisic2012-zy,Brunton22}. Moreover, data-driven approaches leveraging the Koopman operator have gained significant attention due to their ability to learn and model nonlinear dynamics directly from observed data, making them highly applicable in scenarios where explicit system models are difficult to obtain \cite{Mauroy2020-es}.

There have been various approaches to apply machine learning (ML) to wireless communication and networking \cite{Sun19,Qin24,Cao24}. While most existing approaches address problems and challenges in relatively static wireless environments, they may have limitations when applied to highly dynamic environments, such as those in FANETs. Therefore, a different ML approach is needed to address the challenges of highly dynamic FANETs, where the topology evolves over time. In this context, data-driven Koopman approaches are promising solutions for modeling UAV trajectories in FANETs through versatile linearization, enabling more accurate predictions and improving network performance.

 \begin{figure}
    \centering
    \includegraphics[height = 40mm, width=0.5\textwidth]{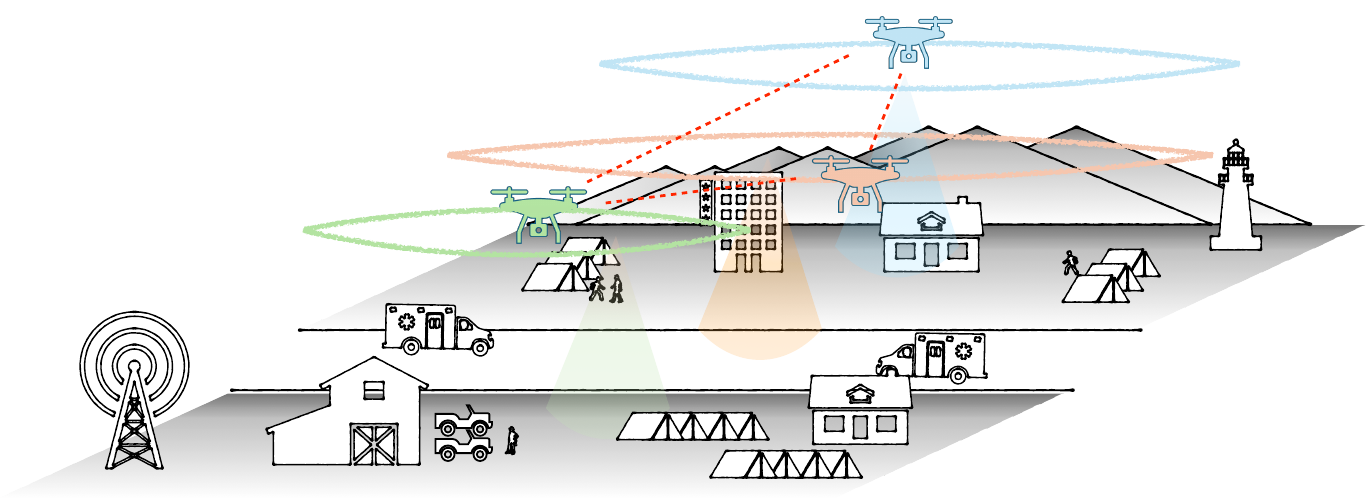}
\caption{A diagram illustrating the deployment of FANETs for surveillance operations, where each UAV is assigned a specific area (denoted by colored circles) to monitor and follows predetermined flight paths. Intra-UAV links (dotted lines) ensure seamless communication and coordination among UAVs,}
    \label{fig:ps}
\end{figure}

In this paper, we explore two approaches to modeling UAV trajectories for FANETs used in surveillance operations, as illustrated in Fig.~\ref{fig:ps}, utilizing the Koopman operator: the distributed approach and the centralized approach\footnote{The centralized approach refers to the use of a global model to manage UAV connectivity, rather than providing an infrastructure such as base stations or access points. In this approach, UAVs still primarily communicate among themselves, except for training data sent to  a central unit for connectivity management.}. In the distributed approach, each UAV models its own dynamics relative to other UAVs and predicts its connectivity with them. In contrast, the centralized approach involves a central unit that trains a model to capture the dynamics of all UAVs. This model is then used to predict future trajectories and connectivity patterns \cite{krishnan2024graph}, which the central unit communicates to the UAVs to inform them of their connectivity. Together, these approaches provide complementary strategies to address the challenges of mobility and connectivity in FANETs.

\subsection{Assumptions}
Throughout this paper, we consider multiple UAVs operating over a designated area for surveillance, with each UAV following an independent flight schedule. The UAVs form an ad hoc network, i.e., FANET, where a UAV can communicate with other UAVs if they are within its communication range. The following additional assumptions are made:
\begin{itemize} 
\item[A1)] In the distributed approach, each UAV operates autonomously to predict its future connectivity with other UAVs by predicting the received \textbf{signal-to-interference-plus-noise ratio} (SINR) from other UAVs over time using the Koopman autoencoder (KAE) \cite{lusch2018deep, azencot2020forecasting}. 
\item[A2)] In the centralized approach, a central unit periodically collects connectivity data from all UAVs in the network. Leveraging the graph KAE (GKAE) \cite{krishnan2024graph}, the central unit performs a one-shot prediction of the SINR between all UAV pairs, providing a global view of the network's connectivity.

\item[A3)] It is assumed that UAVs transmit only their past SINR data to the central unit, rather than their precise coordinates, to preserve the privacy of their location information. 
\end{itemize}

\subsection{Contributions}
The contributions of this work are summarized as follows:
\begin{itemize} 
\item[C1)] In this paper, we explore the application of Koopman-based methods, including the KAE and GKAE, to predict connectivity in FANETs. These methods transform SINR-based channel metrics into a linear latent space, facilitating long-term predictions of network dynamics and connectivity.
\item[C2)] We propose two UAV SINR prediction approaches as follows: the \textbf{centralized approach} where a central unit predicts the received SINR for each UAV, offering a network-wide perspective on connectivity; and the \textbf{distributed Approach} where individual UAVs predict future SINR independently, enabling decentralized decision-making.

\item[C3)] Simulation results demonstrate the effectiveness of our proposed approaches, providing valuable insights into predicting isolation events for the FANET and individual UAVs.
\end{itemize}

\section{Modeling Dynamics via Koopman}
In this section, we provide an overview of the Koopman operator theory \cite{Koopman31} and a data-driven approach for learning nonlinear dynamics (of UAVs) based on this theory. In addition, for graph signals, which represent non-Euclidean data, we present a variant of the data-driven approach tailored to handle such signals.

\subsection{Koopman Operator Theory}

The Koopman operator theory  provides a powerful framework for linearizing the evolution of a nonlinear dynamical system \cite{Koopman31, Budisic2012-zy,Mezi2021,Brunton22}. In this subsection, we present an overview of the  Koopman operator theory.

Denote by $\bx(t)$ the state of a nonlinear dynamical system that evolves over time as follows:
\be
\bx(t+1) = \bF(\bx(t)) \in \cX \subseteq \uR^N, \label{EQ2:xFx}
\ee
where $\bF$ represents the flow map, a nonlinear and unknown function that governs the dynamics of the system. The term $\cX$ denotes the state space, which is a subset of the $N$-dimensional vector space.  
According to the Koopman operator theory, given a observable measurement function, $h: \mathcal{X} \rightarrow \mathbb{R}$, there exists a linear operator called the \emph{Koopman operator},
denoted by $\mathcal{K}$, which can be applied to all such observable $h$, to advance them in time, as follows:
\begin{align}
\cK h = h \circ \bF,
    \label{EQ3:KO}
\end{align}
where $\circ$ represents the composition operator, existing on a smooth manifold. Applying \eqref{EQ3:KO} to \eqref{EQ2:xFx}, we have
\begin{align}
h(\bx(t+1)) = h \circ \bF(\bx(t)) = \cK h(\bx(t)),
    \label{EQ:gg}
\end{align}
where $h(\bx(t))$ is an observable measured at time $t$.
This can be extended to the case with multiple observables. Precisely, let $\bh (t) = [h_1 (t) \ \ldots \ h_M (t)]^\rT$, where $h_m (t) = h_m (\bz(t))$. Then, from \eqref{EQ:gg}, we have
\be 
\bh(t+1) = \cK \bh(t). 
    \label{eq:linear}
\ee 
We can readily show that the Koopman operator is linear, since 
$\cK (c_1 \bh_1 (t) + c_2 \bh_2(t)) = c_1 \bh_1 (t+1) + c_2 \bh_2(t+1)$, where $c_1, c_2 \in \uR$.

If $\bh(t) \in \cH$ and $\cK \bh(t) \in \cH$, where $\cH$ is a finite-dimensional space, $\cH$ becomes a Koopman invariant subspace \cite{Brunton_PLOS}. In this case, the Koopman operator in \eqref{eq:linear} becomes a square matrix, denoted by $\bK$.

\subsection{Koopman Autoencoder}

Although the Koopman operator is useful for linearizing nonlinear dynamical systems, which aids in modeling and prediction, finding the Koopman invariant subspace is challenging. Data-driven techniques, such as dynamic mode decomposition (DMD) methods \cite{Tu14} \cite{Kutz16} and deep learning based approaches \cite{Takeishi17} \cite{lusch2018deep}, have been developed to address this challenge. In particular, the approaches in \cite{Takeishi17} \cite{lusch2018deep} are based on the autoencoder architecture, and called
the KAE, which consists of the  three main components: an encoder, a Koopman matrix, and a decoder, where the encoder and decoder are neural networks. It is designed to estimate the Koopman operator and its invariant subspace, capturing the smooth dynamics through a linear operator.

Let $\cE_{\theta_1} (\bx)$ be the encoder that maps the state vector $\bx$ to a latent variable, denoted by $\bz \in \uR^M$, where $M$ is the dimension of the latent space. Here, $\theta_1$ represents the parameter vector for the encoder. Then, we assume that the encoder performs the linearization so that 
the following relation can hold:
\begin{align} 
\cE_{\theta_1} (\bx(t+1)) & = \bz (t+1) \cr 
& = \bK \bz (t) = \bK \cE_{\theta_1} (\bx(t)),
    \label{EQ:enc}
\end{align}
where $\bK$ is the Koopman matrix. By allowing the inverse mapping with the decoder, denoted by $\cD_{\theta_2} (\bz)$, we can have the original state vector as follows:
\be 
\bx(t) = \cD_{\theta_2} (\bz(t)),
    \label{EQ:dec}
\ee 
where $\theta_2$ is the parameter vector of the decoder. Here, both the encoder and decoder are assumed to be neural networks. 
From \eqref{EQ:enc} and \eqref{EQ:dec}, using the linearity, we can derive the following relation:
\begin{align}
    \bx (t+l) = \cD_{\theta_2} ( \bK^l \cE_{\theta_1} (\bx(t))), \ l = 0, 1,\ldots.  \label{EQ:pred}
\end{align}
Thus, with a given set of state vectors, $\{\bx(0), \ldots, \bx(T)\}$, it might be possible to determine the parameter vectors, $\theta_1$ and $\theta_2$, and Koopman matrix, $\bK$, to minimize the error or loss as follows:
\be 
{\rm Loss}(\theta_1, \theta_2, \bK, L) =\sum_{l=0}^{L-1}
\sum_t ||\bx(t+l) - \cD_{\theta_2} ( \bK^l \cE_{\theta_1} (\bx(t))) ||^2. \label{eq:KAE}
\ee 
\subsection{Graph KAE}
While the traditional KAE is effective in modeling non-linear dynamics using an autoencoder architecture \cite{Takeishi17} \cite{lusch2018deep}, it is typically applied to data originating from a single node, modeled as a Euclidean sequence of input states. This inherently limits its applicability to non-Euclidean data, such as spatio-temporal graphs, where the dynamics arise from interactions among multiple nodes in a graph structure. To extend the modeling capability of the KAE to non-Euclidean sequences, we propose transforming the non-Euclidean graph data into a Euclidean-compatible input. This is achieved by learning a graph embedding that captures the spatial and temporal relationships between nodes at each time step. The graph embedding serves as a compact state vector for each spatio-temporal graph, enabling the KAE to model the dynamics of the sequence effectively while preserving the underlying graph structure and relationships. The operations within the GKAE can be described using the following components:

\subsubsection{Graph Encoder} The graph encoder represents each graph realization as a graph embedding using a graph neural network (GNN), designed to capture latent vectors that preserve the spatial characteristics of the graphs. The latent variable allow us to represent the dynamical system of multiple nodes as an Euclidean sequence of input, enabling the use of KAE to linearize the dynamics of the latent variables, which serve as state vectors for the time-varying graphs. We use a pooling operator for a compact representation as follows:
\be
\bg(t) = \text{POOL}\left(\cE_{\theta_{\text{GNN}}}(\bx(t), \cE(t), \bW(t))\right), \label{eq:pool}
\ee
where \text{POOL$(\cdot$)} represents the average pooling function. 
\subsubsection{KAE} The KAE uses the input graph embedding, which is presumably non-linear due to the non-linearity inherent in the original dynamics and the graph encoder. With processable input of the graph embedding as a the state vector of the dynamical system, we linearize using \eqref{EQ:enc} and the future embedding predictions is found using the Koopman matrix, as seen in \eqref{EQ:pred} and \eqref{EQ:dec}. 

\subsubsection{Graph Decoder } A graph decoder aims at reconstructing from the graph embedding state to the original graph state. Using the input as the graph embedding at any time step, we can reconstruct the node features as 
\be
\hat{\bs}(t+1) = \cD_{\phi_\text{GNN}}(\bg(t+1)), 
\ee
where using the reconstructed node features, the adjacency matrix can be derived based on a predefined rule for edge formation. If this rule is not available, methods such as those in \cite{kipf2016variational} can be employed to reconstruct the edges. In most cases, including this work, we assume that the method for edge formation in the graphs is predefined, which is defined on the graph signals. The operational flow of the GKAE model is given as:
\begin{align}
\notag G(t) &\xrightarrow{\text{POOL}(\mathcal{E}_{\theta_{\text{GNN}}}(\cdot))}  \mathbf{g}(t) \xrightarrow{\mathcal{E}_{\theta_{\text{KAE}}}(\cdot)} \mathbf{h}(t) \\
 &\xrightarrow{\mathbf{K}^p} \mathbf{h}(t+p) \xrightarrow{\mathcal{D}_{\phi_{\text{KAE}}}(\cdot)} \mathbf{g}(t+p) \xrightarrow{\mathcal{D}_{\phi_{\text{GNN}}}(\cdot)} \bx(t+p), \label{eq:flowref}
\end{align}
where $p \ge 1$.

\section{System Model}

In this section, we present the system model and discuss the role of prediction in maintaining connectivity between UAVs in a FANET.

\subsection{Dynamics of UAVs}

Denote by $\cL = \{1, 2, \cdots, L\}$ the set of $L$ UAVs operating within a designated area to perform surveillance tasks. Each UAV is assumed to follow a deterministic dynamic trajectory, which may vary depending on the specific characteristics of the UAV. For simplicity, we only consider the two-dimensional (2D) movement of UAVs. The  location of UAV $l$ at time $t$ is denoted by $\bx_l(t) \in \cX \subseteq \uR^2$, where $\cX$ is the area of operation.

The dynamics for the $l$th UAV \cite{shao2023path} is given by
\begin{align}
    x_{l}(t+1) &= x_l(t) + u_{l} \cos{\psi_l} + v_{iw} \cos{\theta_{iw}}\\ 
    y_{l}(t+1) &= y_l(t) + u_{l} \sin{\psi_l} + v_{iw} \sin{\theta_{iw}}\\
    \psi_l(t+1) &= \psi_l(t) + r_l,
\end{align}
where $\bx_l(t) = [x_l(t) \  x_{l}(t)]^\top$ are the x and y coordinates of UAV $l$ at time $t$. Here, $u_l$, $\psi_l$ and $r_l$ 
represent the forward velocity, turning rate and heading angle, respectively. Furthermore, the wind velocity and wind angle are denoted by $v_{iw}$ and $\theta_{iw}$, respectively. 

\subsection{Inter-UAV Communication}

We assume that each UAV is equipped with a transceiver and is able to communicate with other UAVs. The UAV is assumed to have a unique beacon signal for identification, which is transmitted periodically. Let $P_l$ be the transmit power of UAV~$l$. The SINR from UAV $j$ to UAV $i$ at time $t$ is given as 
\begin{align}
\gamma_{i \leftarrow j}(t) =
\frac{P_j d_{i, j}(t)^{-\eta}}{N_0 B + \sum_{k \ne j} P_k d_{i, k}(t)^{-\eta}} ,
    \label{EQ:SINR}
\end{align}
where $d_{i, j} (t) = ||\bx_i (t) - \bx_j (t) ||$, $N_0$, $B$ and $\eta$ represent the Euclidean distance, noise spectral density, bandwidth and the path-loss exponent, respectively. 
At each time $t$, the SINRs at UAV $i$  from all the other UAVs with $j \in \cL \backslash \{i\}$ can be written~as 
\be
\boldsymbol{\gamma}_i(t) = [\gamma_{i \leftarrow 1}(t) \  \cdots \ \gamma_{i \leftarrow i-1}(t) 
\  \gamma_{i \leftarrow i+1}(t),\ \cdots, \gamma_{i \leftarrow L}(t)]^\top .
\ee

The index set of neighbors of the UAV is consequently defined for a communication SINR threshold of $\kappa >0$, which is given by 
\be
\cN_i(t) = \{j:\gamma_{i \leftarrow j}(t) \ge \kappa\}.
\ee

\subsection{Inter-UAV Connectivity}

In a FANET, the connectivity of UAVs is influenced not simply by their proximity but by their SINRs. To quantitatively identify the connectivity, we focus on the event of isolation characterized by SINR outage as follows.
At time $t$, a UAV is considered isolated when it fails to establish a reliable communication link with any other UAVs, resulting in an empty set of neighbors to communicate. Accordingly, the event of isolation for UAV $l$ is defined as 
\be
\cI_l(t) = \{\cN_l(t) = \emptyset\},
\ee
while the network isolation event is defined as 
\begin{equation}
\mathcal{I}_\text{network}(t) = \bigvee_{l \in \mathcal{L}} \{\mathcal{N}_l(t) = \emptyset\},
\end{equation}
where $\bigvee$ represents the disjunction operator.

Note that in hostile environments, UAVs may experience unexpected connectivity loss due to attacks. Unlike these abrupt, attack-induced connectivity disruptions, isolation events can be predicted based on the correlated patterns of SINRs. By leveraging these isolation predictions, UAVs can proactively forward their data packets to other UAVs, thereby maintaining reliable inter-UAV communication. However, predicting isolation in a FANET is challenging due to the time-varying SINRs and spatial interactions among UAVs through interference, i.e., $\sum_{k \ne j} P_k d_{i, k}(t)^{-\eta}$ in the denominator of \eqref{EQ:SINR}. To address this challenge, we will introduce GKAE based isolation prediction algorithms in the next section.


\section{Proposed Approaches}
\begin{figure}
    \centering
    \includegraphics[height = 50mm, width=0.5\textwidth]{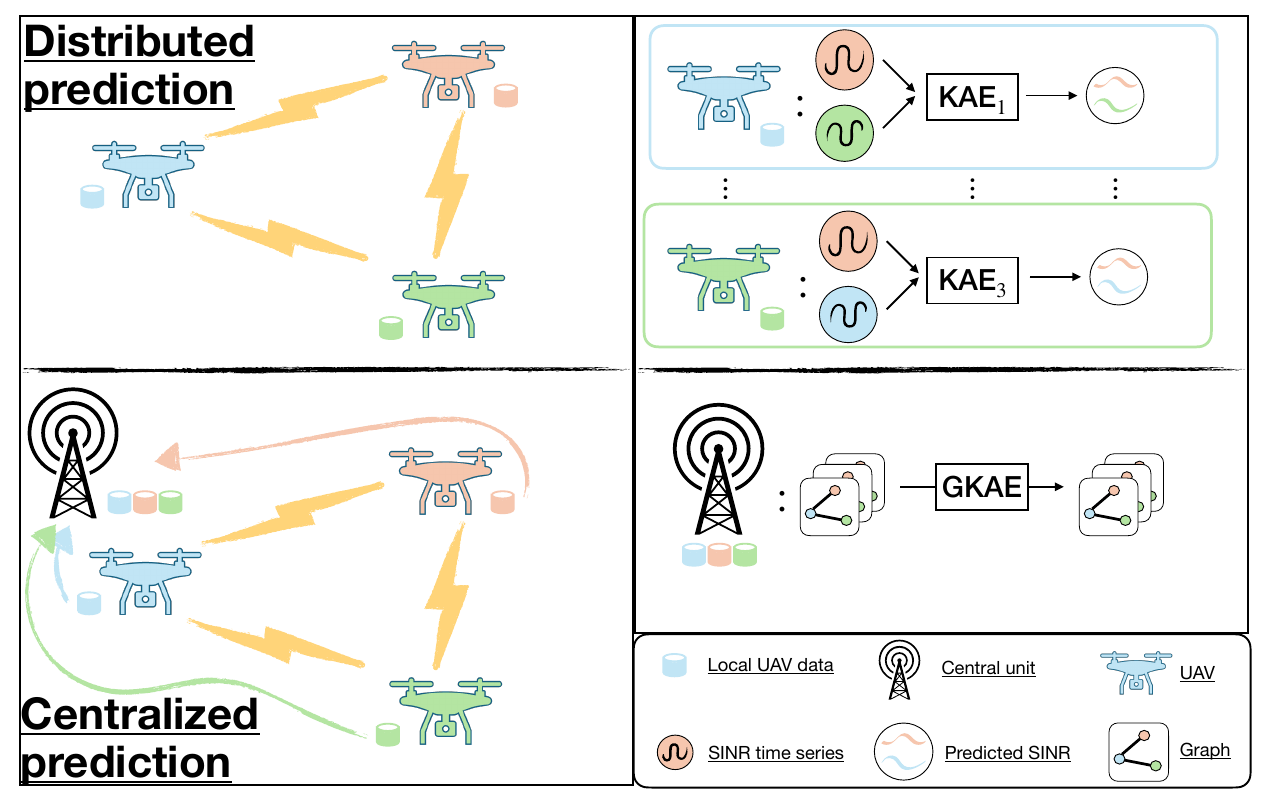}
\caption{Two approaches are proposed for proactively detecting isolation events: 1) In the distributed approach, each UAV utilizes its local observations to train a KAE, enabling it to predict connectivity with other UAVs. 2) In the centralized approach, a central unit gathers information from all UAVs within the operational area to collectively assess the connectivity of each UAV.}
    \label{fig:2}
\end{figure}

In this section, as Fig.~\ref{fig:2} illustrates, we elaborate on the two approaches for predicting the SINR and, consequently, the events of isolation.
The centralized approach relies on a central unit with global knowledge of the SINR values for all UAVs within the operational area. In contrast, the distributed approach assumes that each UAV operates independently, using only its local information without access to the global network view.

\subsection{Centralized Approach}
We assume that the central unit is equipped with a GKAE model. Utilizing the collected data, which captures the SINR values for each UAV pair, the model can predict the future SINR values for each UAV pair. We firstly translate each time step $t$ for the FANET as a time-varying graph. 
Let $G(t) = (\mathcal{L}, \mathcal{E}(t), \mathbf{S}(t), \mathbf{W}(t))$ represent the quadruple for a time-varying graph, where:
\begin{itemize}
    \item $\mathcal{L}$: Node set of UAVs, with cardinality $|\mathcal{L}| = N$ (number of UAVs).
    \item $\mathcal{E}(t)$: Edge set at time $t$, with $|\mathcal{E}(t)|$ representing the number of edges, which is given as
    \be
    \cE(t) = \{(i, j) : \gamma_{i \leftarrow j} \ge \kappa, i \ne j \in \cL\} . \label{eq:setedge}
    \ee
    \item $\mathbf{S}(t) \in \mathbb{R}^{L \times (L-1)} = [\boldsymbol{\gamma}_1(t)^\top\ \boldsymbol{\gamma}_2(t)^\top \ldots \boldsymbol{\gamma}_L(t)^\top]^\top$: Node feature matrix at time $t$, where $L-1$ represents the dimensionality per node, which is the SINR received from each node. 
    \item $\mathbf{W}(t) \in \mathbb{R}^{L \times L}$: Weighted adjacency matrix at time $t$, where $\mathbf{W}_{i,j}(t)$ represents the weight of the edge between node $i$ and node $j$, which is the Euclidean distance between $i$ and $j$
    \begin{align}
    \mathbf{W}_{i,j}(t) =
    \begin{cases}
    d_{i,j}(t), & \text{if } j \in \mathcal{N}_i(t) \text{ and } i \neq j, \\
    0, & \text{otherwise.} \label{eq:setweight}
    \end{cases}
    \end{align}
\end{itemize}

We use graph embedding methods \cite{xu2021understanding} as means of converting the non-Euclidean graph data as a graph embedding $\bg(t) \in \mathbb{R}^b$ which represents the graph as a latent variable while restoring its spatial characteristics. The graph embedding can be found using encoder architectures which consists of GNN layers, while the graph decoder which takes as input the graph embedding can be used for reconstructing the graph node features, which can be used for subsequently reconstructing the edge set and the adjacency matrix using \eqref{eq:setedge} and \eqref{eq:setweight}, respectively. We have described a \emph{vanilla} graph autoencoder (GAE) that takes an input and reconstructs the same graph, which, in the case of a time-varying graph, is restricted to the same time step $t$. 

We enable a dynamical autoencoder by incorporating the KAE between the graph encoder and the graph decoder to model the non-linear dynamics in the graph embedding as a linear model via the Koopman matrix. The GKAE acts as a dynamical autoencoder, takes as input $G(t)$ and outputs $G(t+p)$, which represents the graph $p$ prediction steps later, as shown in \eqref{eq:flowref}. Using the predictions, we can estimate the isolation events as described in \eqref{fig:iso}. This is essential for ensuring network connectivity, enabling the delivery of priority communication tasks without interruption. By proactively addressing the estimated isolation events, delayed reactions can be avoided, thereby maintaining the efficiency and reliability of the surveillance operation.

The centralized approach enables globally optimal solutions for tasks such as resource allocation, routing, and trajectory planning. It can also facilitate covert communication for ground users under UAV surveillance \cite{krishnan2024graph}. However, this approach requires a stable communication link between the UAVs and the central unit, which may not always be feasible in dynamic environments. Moreover, processing data for multi-UAV networks often involves graph-based techniques, resulting in higher computational overhead. Motivated by this, we additionally present its distributed version in the following subsection.

\subsection{Distributed Approach}
The assumption of the existence of a central unit may not hold in uncontrollable or dynamic environments. Moreover, the communication link between the central unit and the UAVs is susceptible to disconnections, which can result in interruptions in data transmission, delays in decision-making, and a loss of coordination across the UAV network. These vulnerabilities can significantly undermine the effectiveness of surveillance operations, underscoring the importance of developing robust and adaptive network architectures that do not rely solely on centralized control. In the distributed method, we deploy a KAE on each UAV. Using past SINR data, the KAE can predict future neighboring UAVs, enabling the UAV to schedule priority communications effectively. This proactive approach also allows the UAV to adjust its transmit power to prevent isolation events, ensuring seamless connectivity and improved network performance. Using the local-decision making by training $L$ UAVs, we aim to provide a similar performance comparable to that of the centralized approach. By optimizing their local models and actions, the UAVs collectively emulate the global coordination typically achieved through centralized methods, ensuring robust and scalable performance even in dynamic or uncontrollable environments.

Unlike the centralized approach, the distributed approach allows UAVs to operate autonomously using localized data. This method is highly scalable, especially as the number of UAVs increases. However, the decisions made are inherently suboptimal, as they are based on individual UAV perspectives rather than a global network view. Both the pros and cons of the centralized and distributed approaches will be compared by simulation in the next section.


\section{Simulation Results}

\subsection{FANET Settings}

\begin{figure}
    \centering
    \includegraphics[height = 40mm, width=0.5\textwidth]{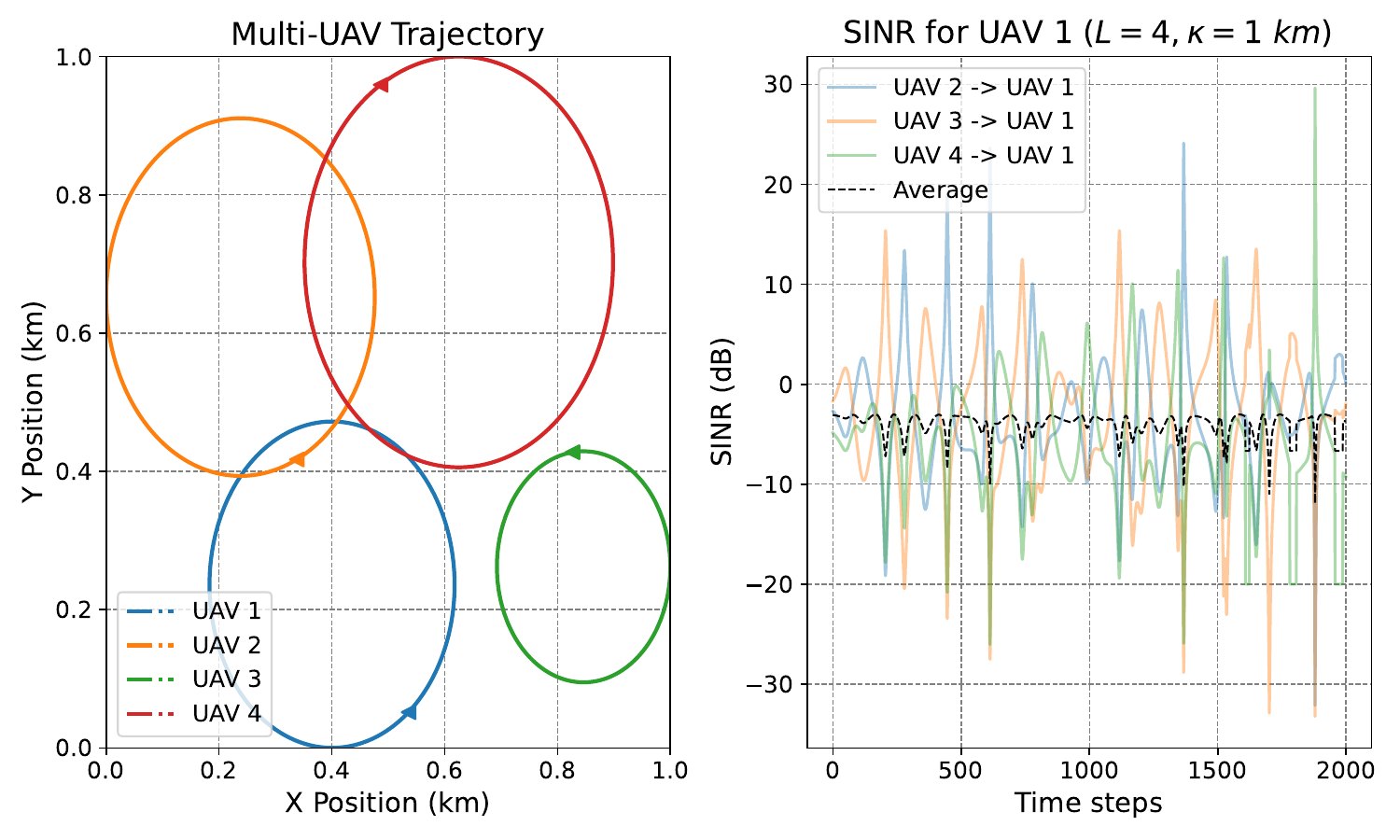}
\caption{(Left) The dynamics of $L = 4$ UAVs with varying cycle durations; (right) Quasi-periodic SINR for UAV $1$ over $2000$ time steps.}
    \label{fig:3}
\end{figure}

It is assumed that a FANET consisting of multiple UAVs is used for surveillance, where UAVs communicate with each other to exchange data. In particular, we simulate the dynamics of $L$ fixed-wing UAVs \cite{shao2023path}. The UAVs are deployed over an operational area to maximize target coverage, minimize non-overlapping regions, and avoid collisions. Each UAV follows a predetermined trajectory designed for periodic surveillance.
The distinct cycle duration of each UAV is determined by the total area under surveillance and is, in terms of dynamics, directly influenced by the turning rate $\psi_i$ of each UAV. Due to the varying turning rates $\psi_i$, each UAV exhibits a different cycle duration and periodicity. Consequently, the SINR curves for each UAV display quasi-periodic behavior, which becomes predictable, as illustrated in Fig.~\ref{fig:3}.

To simulate variations in surveillance responsibilities, UAVs covering larger areas are assigned turning rates sampled from a uniform distribution: $\psi_i \sim \cU(\psi_{\text{min}}, \psi_{\text{max}})$, for which the parameters are summarized in Table~\ref{tab:params}.
\begin{table}[h!]
    \centering
    \caption{Simulation parameters}
    \begin{tabular}{|c|c|}
        \hline
        \hline
        \multicolumn{2}{|c|}{\textbf{Model architecture}} \\
        \hline
        \hline
        Number of layers (centralized) & 15\\
        Number of layers (distributed) & 11\\
        \hline
        \hline
        \multicolumn{2}{|c|}{\textbf{UAV dynamics}} \\
        \hline
        \hline  
        Area of operation & 1000 $\times$ 1000 m$^2$\\
        Number of UAVs & 4\\
        Velocity & $\cU(10, 15)$ m/s \\
        Wind velocity & $10^{-8}$ m/s\\
        Turning rate &  $\cU(0.01, 0.05)$ rad.\\
        Wind direction & $10^{-8}$ rad.\\
        \hline
        \hline        
        \multicolumn{2}{|c|}{\textbf{Channel model}} \\
        \hline
        \hline
        Communication radius of UAVs $\kappa$ & 500 m\\
        Transmit power & 0.1 W \\
        Path loss & 2 \\
        Noise spectral density & -174 dBm/Hz \\
        Bandwidth & 1 MHz \\
        \hline
        \hline
    \end{tabular}
    \label{tab:params}
\end{table}


\subsection{Performance Metrics}
In the centralized approach, SINR predictions for the entire FANET are denoted as $\hat{\bS}(t)$, whereas in the distributed approach, predictions are made per UAV and represented as$\hat{\bs}_i(t)$. 

Given the initial input at $t = 0$ as $\bs_i(0) \in \mathbb{R}^{L-1}$ for the distributed approach or $\bS(0) \in \mathbb{R}^{L \times L-1}$ for the centralized approach, the total predictive error over $P$ prediction steps is defined as:
\begin{align}
\epsilon(P) =  \frac{1}{P} \sum_{t = 1}^P \sum_{l = 1}^L \left\|\bs_l(t) - \hat{\bs}_l(t)\right\|^2,
\end{align}
where $\bs_l(t)$ denotes the ground truth SINR for UAV $l$, and $\hat{\bs}_l(t)$ represents the predicted SINR. 

The performance metric $\epsilon(P)$ can thus be evaluated for both approaches:
\begin{itemize}
    \item In the \textbf{centralized approach}, $\hat{\bs}_l(t)$ corresponds to individual rows of $\hat{\bS}(t)$, where all UAVs' SINR values are jointly predicted and compared against the ground truth.
    \item In the \textbf{distributed approach}, $\hat{\bs}_l(t)$ represents the SINR predictions generated independently for each UAV $l$.
\end{itemize} 
Using the predicted SINR, we evaluate how well the model can estimate the network isolation event (centralized approach) or the isolation of a single UAV (distributed approach). Let $\hat{\cI}_{\text{network}}(t)$ and $\hat{\cI}_{{l}}(t)$ represent the predicted isolation events using the predicted SINR $\hat{\bS}(t)$ and $\hat{\bs}_l(t)$ using the centralized and the distributed approaches, respectively. The event classifications are described in Table~\ref{tab:isolation_classification}. 

\begin{table}[h!]
\caption{Classification of Isolation Events for UAVs or Network}
    \centering
    \begin{tabular}{|c||c|c|}
        \hline
        \textbf{Classification} & \textbf{Actual Event} & \textbf{Predicted Event} \\ \hline
        True Positive (TP)      & Isolation  & Isolation \\ \hline
        False Positive (FP)     & No Isolation  & Isolation \\ \hline
        True Negative (TN)      & No Isolation  & No Isolation \\ \hline
        False Negative (FN)     & Isolation  & No Isolation \\ \hline
    \end{tabular}
    \label{tab:isolation_classification}
\end{table}
Over $P$ steps of predicted future SINR, the F1-score is defined as
\be
\text{F1-score} = \frac{2 \cdot \text{TP}}{2 \cdot \text{TP} + \text{FP} + \text{FN}} ,
\ee
while the false alarm rate (FAR) is given by
\be
\frac{\text{FP}}{\text{FP + TN}}.
\ee

\subsection{Results}
\subsubsection{Predictive Performance} 
\begin{figure}[t]
    \centering
    \includegraphics[width=\linewidth]{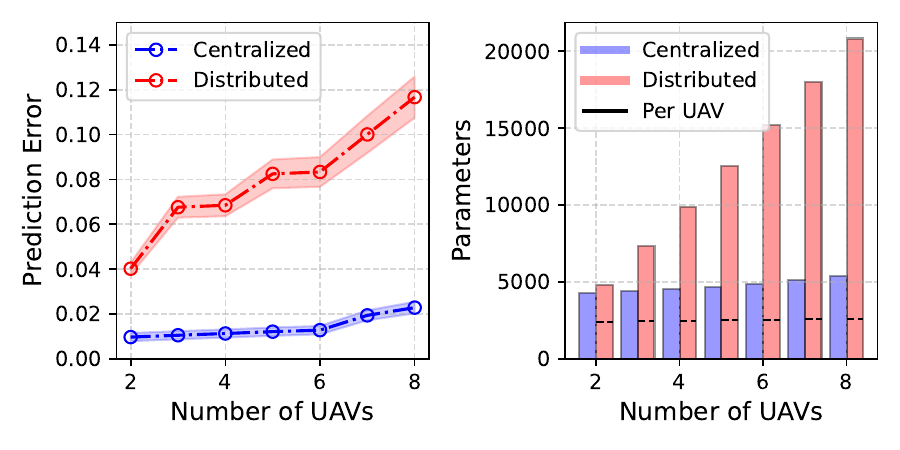}
\caption{(Left) Prediction error over varying UAVs when $P = 50$ time steps, aggregated over 1900 trajectory predictions, each starting from a unique initial point. (Right) Computational complexity comparison between training the GKAE and the KAE approaches, where the per UAV computational cost is highlighted using a black line. }
    \label{fig:pe}
\end{figure}

In Fig.~\ref{fig:pe}, we compare the prediction error and computational complexity, measured by the total trainable parameters, for both the centralized and distributed approaches. The centralized approach employs 15 layers, including 4 layers dedicated to the graph encoder and decoder. In contrast, the distributed approach relies solely on the KAE, comprising 11 layers. In the centralized approach, the central unit has access to a global view of the UAV network, enabling it to account for all interactions among UAVs comprehensively. In contrast, the distributed approach relies solely on local information, specifically the SINR curves of individual UAVs. This fundamental difference explains the growing disparity in prediction error between the two approaches as the number of UAVs increases. With a higher number of UAVs, the interactions within the network become more complex, and the distributed approach loses significant interaction-related information, leading to greater predictive error. Conversely, the centralized approach benefits from its holistic view, enabling more accurate predictions, resulting in an average improvement of $75- 80\%$. However, this accuracy comes at the cost of computational complexity. Training the centralized model is resource-intensive, requiring approximately $79 - 107\%$ more trainable parameters compared to the distributed approach per UAV. This trade-off highlights the balance between computational efficiency and prediction accuracy when choosing between centralized and distributed methods.
\begin{figure}[h!]
    \centering
    \includegraphics[width=\linewidth]{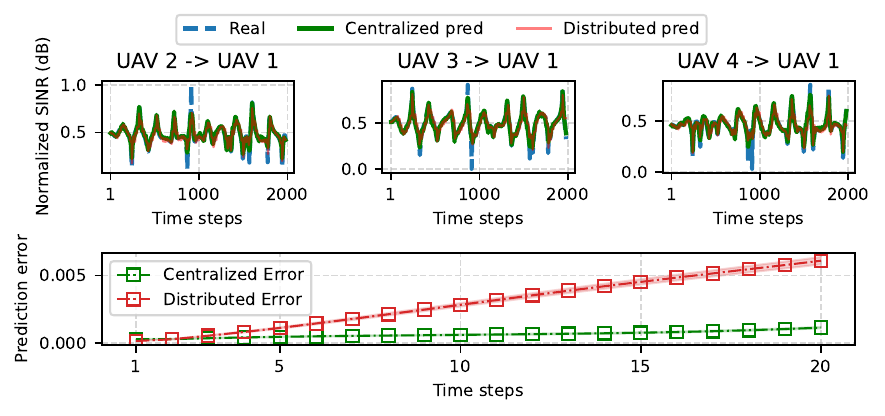}
    \caption{(Top) Comparison of 20-step predictions up to 2000 time steps for UAV 1 using two different approaches. (Bottom) Aggregated prediction error comparison for the two approaches over 1000 20-step predictions.}
    \label{fig:ds}
\end{figure}

In Fig.~\ref{fig:ds}, we illustrate the 20-step predicted SINR curves for UAV 1. The results demonstrate a clear linear increase in prediction error over time for both the centralized and distributed approaches. However, it is evident that the centralized predictions consistently exhibit lower prediction errors compared to the distributed predictions.

\subsubsection{Estimation of Isolation Events}
\begin{figure}
    \centering
    \includegraphics[width=\linewidth]{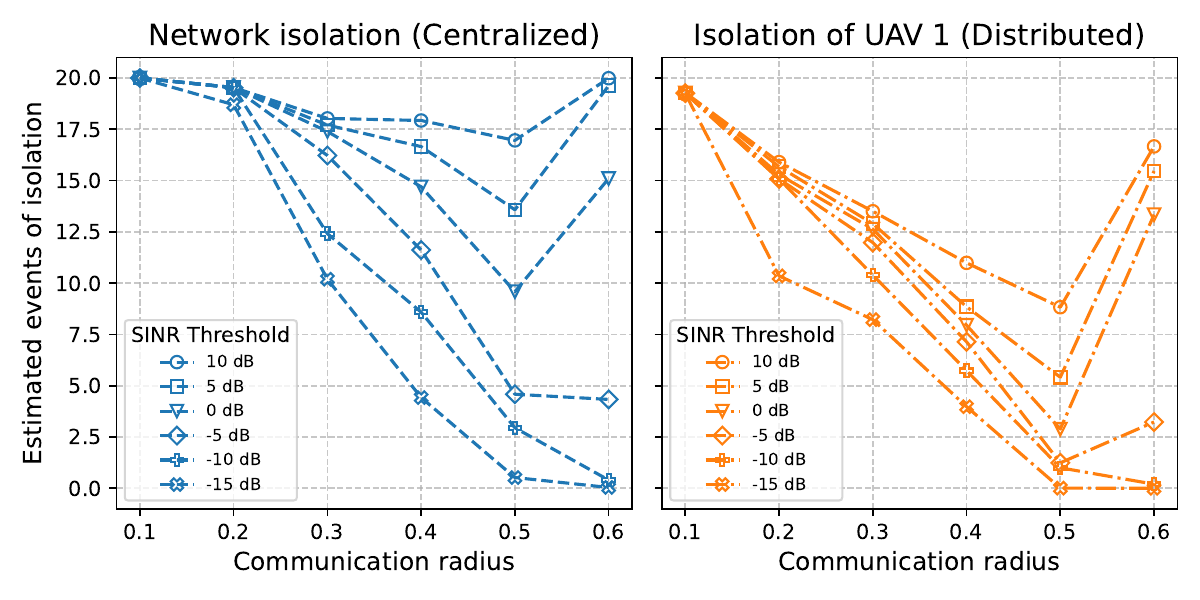}
\caption{Evaluating the network isolation (left) and the isolation of a single UAV (right) over varying communication radius and different SINR thresholds for isolation.}
    \label{fig:iso}
\end{figure}

Some FANET operations require a consistently high SINR to ensure robust communication and minimal interference, particularly in scenarios demanding precise coordination. Conversely, other operations can tolerate lower SINR thresholds, prioritizing broader connectivity and flexibility over stringent signal quality requirements. In Fig.~\ref{fig:iso}, we compare the total estimated network isolation events with the isolation events of a single UAV, indexed at $l = 1 $. The total network isolation events consistently exceed those of a single UAV, reflecting the subset relationship. As the communication radius increases, more UAVs fall within the range of each other, leading to a higher number of potential connections. While a larger communication radius might intuitively seem beneficial for maintaining connectivity, it comes with significant trade-offs. Each UAV communicates using specific transmission power, and as more UAVs enter the communication radius, the combined interference from their transmissions intensifies. Achieving higher SINR thresholds ($\ge 0 \ \text{dB}$), is increasingly challenging because the signal strength of a UAV must now dominate over the cumulative interference from nearby UAVs. In essence, while a larger communication radius facilitates connectivity by bringing more UAVs within range, it also exacerbates interference, creating a trade-off between extending communication range and maintaining high-quality links.

\begin{figure}
    \centering
    \includegraphics[width=\linewidth]{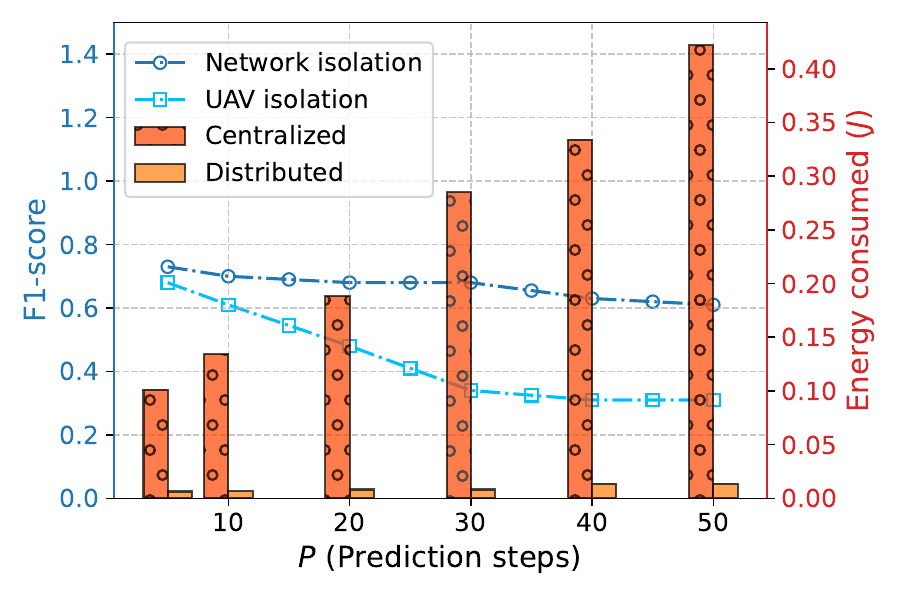}
\caption{Comparison of F1-score (Y1 axis) for different approaches across varying prediction steps, alongside the average energy consumption; (Y2 axis) in Joules for predictions over $P$ steps.}
    \label{fig:f1}
\end{figure}
In Fig.~\ref{fig:f1}, we compare the F1-scores of the proposed approaches for detecting network isolation and individual UAV isolation. Accurate estimation of these events heavily relies on precise SINR curve predictions, which, as seen in Fig.~\ref{fig:pe}, are sub-optimal in the distributed predictions compared to the centralized approach. On average, the centralized approach achieves a $7.35\% - 96.77\%$ improvement in F1-score as the number of prediction steps increases, compared to detecting UAV isolation events using distributed predictions. On the other hand, upon deployment of the trained models for predictions, the distributed approach uses $94.05\% - 96.92\%$ lower energy for making predictions. 
\begin{figure}
    \centering
    \includegraphics[width=0.4\textwidth]{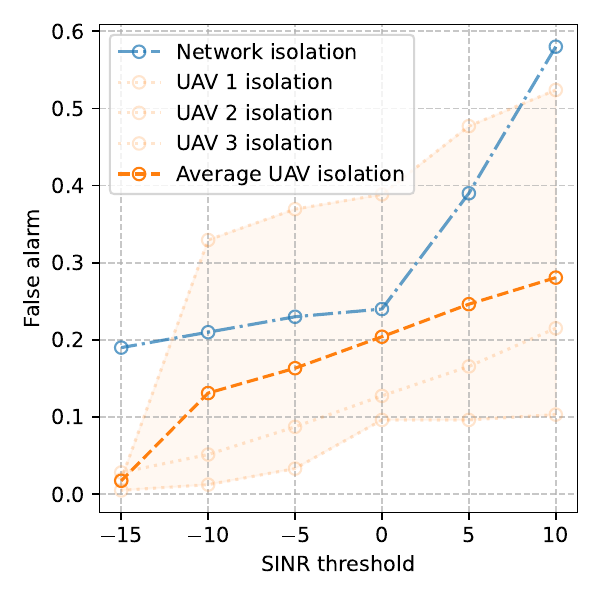}
\caption{Comparison of false alarm rates between the centralized and distributed approaches for detecting isolation events}
    \label{fig:fa}
\end{figure}
When deploying the solutions, it is crucial to avoid false alarms of isolation, as these can lead to unnecessary interventions, resource wastage, and potential disruptions in system operations. In Fig.~\ref{fig:fa}, we compare the false alarm rates across varying SINR thresholds in a FANET with $L = 3$ UAVs. As the SINR thresholds increase, particularly for UAVs with a larger communication radius, the number of isolation events rises, which leads to a higher number of isolation events (TP) while also increases the risk of false alarms due to stricter SINR requirements. On average, network isolation events exceed those of individual UAVs because the network isolation encompasses the subset of all UAV isolation events. However, predicting SINR curves for UAVs using only local information poses challenges, as evidenced by UAV 3, which shows a disproportionately high false alarm rate compared to UAV 1 and UAV 2.


\section{Conclusion}
This paper studied a novel method for predicting the connectivity of UAVs in FANETs using neural networks and Koopman theory. Two distinct approaches for training the model were proposed: centralized and distributed. The centralized approach leveraged global network data to achieve higher SINR prediction accuracy, while the distributed approach enabled scalable and privacy-preserving SINR predictions by individual UAVs. Simulation results validated the effectiveness of both methods in detecting isolation events, demonstrating their potential to enhance routing strategies and enable energy-efficient communication. Future work will focus on improving distributed predictions by periodically exchanging models to better approximate global knowledge.

\bibliographystyle{ieeetr}
\bibliography{Koopman}

\begin{thebibliography}{10}

\bibitem{Toh101}
C.~Toh, {\em Ad Hoc Mobile Wireless Networks: Protocols and Systems}.
\newblock Pearson Education, 2001.

\bibitem{Loo16e}
J.~Loo, J.~L. Mauri, and J.~H. Ortiz, eds., {\em Mobile Ad Hoc Networks: Current Status and Future Trends}.
\newblock CRC Press, 2016.

\bibitem{Khan17}
M.~A. Khan, A.~Safi, I.~M. Qureshi, and I.~U. Khan, ``Flying ad-hoc networks (fanets): A review of communication architectures, and routing protocols,'' in {\em 2017 First International Conference on Latest trends in Electrical Engineering and Computing Technologies (INTELLECT)}, pp.~1--9, 2017.

\bibitem{Nemati2022-sb}
M.~Nemati, B.~Al~Homssi, S.~Krishnan, J.~Park, S.~W. Loke, and J.~Choi, ``Non-terrestrial networks with {UAVs}: A projection on flying ad-hoc networks,'' {\em Drones}, vol.~6, p.~334, Oct. 2022.

\bibitem{Cicek19}
C.~T. Cicek, H.~Gultekin, B.~Tavli, and H.~Yanikomeroglu, ``{UAV} base station location optimization for next generation wireless networks: Overview and future research directions,'' in {\em 2019 1st International Conference on Unmanned Vehicle Systems-Oman (UVS)}, pp.~1--6, 2019.

\bibitem{He24}
H.~He, W.~Yuan, S.~Chen, X.~Jiang, F.~Yang, and J.~Yang, ``Deep reinforcement learning-based distributed {3D} {UAV} trajectory design,'' {\em IEEE Transactions on Communications}, vol.~72, no.~6, pp.~3736--3751, 2024.

\bibitem{Koopman31}
B.~O. Koopman, ``Hamiltonian systems and transformation in {H}ilbert space,'' {\em Proceedings of the National Academy of Sciences}, vol.~17, no.~5, pp.~315--318, 1931.

\bibitem{Budisic2012-zy}
M.~Budisi{\'c}, R.~Mohr, and I.~Mezi{\'c}, ``Applied {K}oopmanism,'' {\em Chaos}, vol.~22, p.~047510, Dec. 2012.

\bibitem{Brunton22}
S.~L. Brunton, M.~Budi\v{s}i\'{c}, E.~Kaiser, and J.~N. Kutz, ``Modern {K}oopman theory for dynamical systems,'' {\em SIAM Review}, vol.~64, no.~2, pp.~229--340, 2022.

\bibitem{Mauroy2020-es}
A.~Mauroy, I.~Mezi{\'c}, and Y.~Susuki, eds., {\em The {K}oopman Operator in Systems and Control}.
\newblock Lecture notes in control and information sciences, Cham, Switzerland: Springer Nature, 1~ed., Feb. 2020.

\bibitem{Sun19}
Y.~Sun, M.~Peng, Y.~Zhou, Y.~Huang, and S.~Mao, ``Application of machine learning in wireless networks: Key techniques and open issues,'' {\em IEEE Communications Surveys \& Tutorials}, vol.~21, no.~4, pp.~3072--3108, 2019.

\bibitem{Qin24}
Z.~Qin, L.~Liang, Z.~Wang, S.~Jin, X.~Tao, W.~Tong, and G.~Y. Li, ``{AI} empowered wireless communications: From bits to semantics,'' {\em Proceedings of the IEEE}, vol.~112, no.~7, pp.~621--652, 2024.

\bibitem{Cao24}
X.~Cao, B.~Yang, K.~Wang, X.~Li, Z.~Yu, C.~Yuen, Y.~Zhang, and Z.~Han, ``Ai-empowered multiple access for {6G}: A survey of spectrum sensing, protocol designs, and optimizations,'' {\em Proceedings of the IEEE}, vol.~112, no.~9, pp.~1264--1302, 2024.

\bibitem{krishnan2024graph}
S.~Krishnan, J.~Park, G.~Sherman, B.~Campbell, and J.~Choi, ``Graph {K}oopman autoencoder for predictive covert communication against {UAV} surveillance,'' in {\em 2024 IEEE 99th Vehicular Technology Conference (VTC2024-Spring)}, pp.~1--5, 2024.

\bibitem{lusch2018deep}
B.~Lusch, J.~N. Kutz, and S.~L. Brunton, ``Deep learning for universal linear embeddings of nonlinear dynamics,'' {\em Nature communications}, vol.~9, no.~1, p.~4950, 2018.

\bibitem{azencot2020forecasting}
O.~Azencot, N.~B. Erichson, V.~Lin, and M.~Mahoney, ``Forecasting sequential data using consistent {K}oopman autoencoders,'' in {\em International Conference on Machine Learning}, pp.~475--485, PMLR, 2020.

\bibitem{Mezi2021}
I.~Mezić, ``{K}oopman operator, geometry, and learning of dynamical systems,'' {\em Notices of the American Mathematical Society}, vol.~68, pp.~1087--1105, Aug. 2021.

\bibitem{Brunton_PLOS}
S.~L. Brunton, B.~W. Brunton, J.~L. Proctor, and J.~N. Kutz, ``{K}oopman invariant subspaces and finite linear representations of nonlinear dynamical systems for control,'' {\em PLOS ONE}, vol.~11, pp.~1--19, 02 2016.

\bibitem{Tu14}
J.~H. Tu, C.~W. Rowley, D.~M. Luchtenburg, S.~L. Brunton, and J.~N. Kutz, ``On dynamic mode decomposition: Theory and applications,'' {\em Journal of Computational Dynamics}, vol.~1, no.~2, pp.~391--421, 2014.

\bibitem{Kutz16}
J.~N. Kutz, S.~L. Brunton, B.~W. Brunton, and J.~L. Proctor, {\em Dynamic Mode Decomposition}.
\newblock Philadelphia, PA: Society for Industrial and Applied Mathematics, 2016.

\bibitem{Takeishi17}
N.~Takeishi, Y.~Kawahara, and T.~Yairi, ``Learning {K}oopman invariant subspaces for dynamic mode decomposition,'' in {\em Proceedings of the 31st International Conference on Neural Information Processing Systems}, NIPS'17, (Red Hook, NY, USA), p.~1130–1140, Curran Associates Inc., 2017.

\bibitem{kipf2016variational}
T.~N. Kipf and M.~Welling, ``Variational graph auto-encoders,'' {\em arXiv preprint arXiv:1611.07308}, 2016.

\bibitem{shao2023path}
X.~Shao, H.~Liu, W.~Zhang, J.~Zhao, and Q.~Zhang, ``Path driven formation-containment control of multiple uavs: A path-following framework,'' {\em Aerospace Science and Technology}, vol.~135, p.~108168, 2023.

\bibitem{xu2021understanding}
M.~Xu, ``Understanding graph embedding methods and their applications,'' {\em SIAM Review}, vol.~63, no.~4, pp.~825--853, 2021.

\end{thebibliography}
\end{document}